\documentclass[runningheads]{llncs}
\usepackage{graphicx}

\usepackage{tikz}
\usepackage{comment}
\usepackage{amsmath,amssymb} 
\usepackage{color}

\usepackage{subcaption}
\usepackage{cite}
\usepackage{gensymb}

\newcommand{\etal}{et al.}
\newcommand{\eg}{e.g.}
\newcommand{\ie}{i.e.}

\newcommand{\repeatthanks}{\textsuperscript{\thefootnote}}

\begin{document}
\pagestyle{headings}
\mainmatter
\def\ECCVSubNumber{1276}  

\title{MessyTable: Instance Association in \\ Multiple Camera Views}


\titlerunning{MessyTable}
%
\author{Zhongang Cai \inst{1}\thanks{indicates equal contribution.} \and
Junzhe Zhang \inst{1,2}\repeatthanks \and
Daxuan Ren\inst{1,2} \and
Cunjun Yu\inst{1} \and
\\Haiyu Zhao\inst{1}\and
Shuai Yi\inst{1}\and
Chai Kiat Yeo\inst{2} \and
Chen Change Loy\inst{2}}
\authorrunning{Cai et al.}
%
\institute{SenseTime Research
\email{\{caizhongang,yucunjun,zhaohaiyu,yishuai\}@sensetime.com}
\and Nanyang Technological University
\email{\{junzhe001,daxuan001\}@e.ntu.edu.sg,\{asckyeo,ccloy\}@ntu.edu.sg}}
\maketitle


\begin{abstract}

We present an interesting and challenging dataset that features a large number of scenes with messy tables captured from multiple camera views. Each scene in this dataset is highly complex, containing multiple object instances that could be identical, stacked and occluded by other instances.
The key challenge is to associate all instances given the RGB image of all views. The seemingly simple task surprisingly fails many popular methods or heuristics that we assume good performance in object association.
The dataset challenges existing methods in mining subtle appearance differences, reasoning based on contexts, and fusing appearance with geometric cues for establishing an association.
We report interesting findings with some popular baselines, and discuss how this dataset could help inspire new problems and catalyse more robust formulations to tackle real-world instance association problems.\footnote{project page: \textcolor{magenta}{\url{https://caizhongang.github.io/projects/MessyTable/}}.}

\end{abstract}

\begin{figure}[t]
  \centering
  \includegraphics[width=\textwidth]{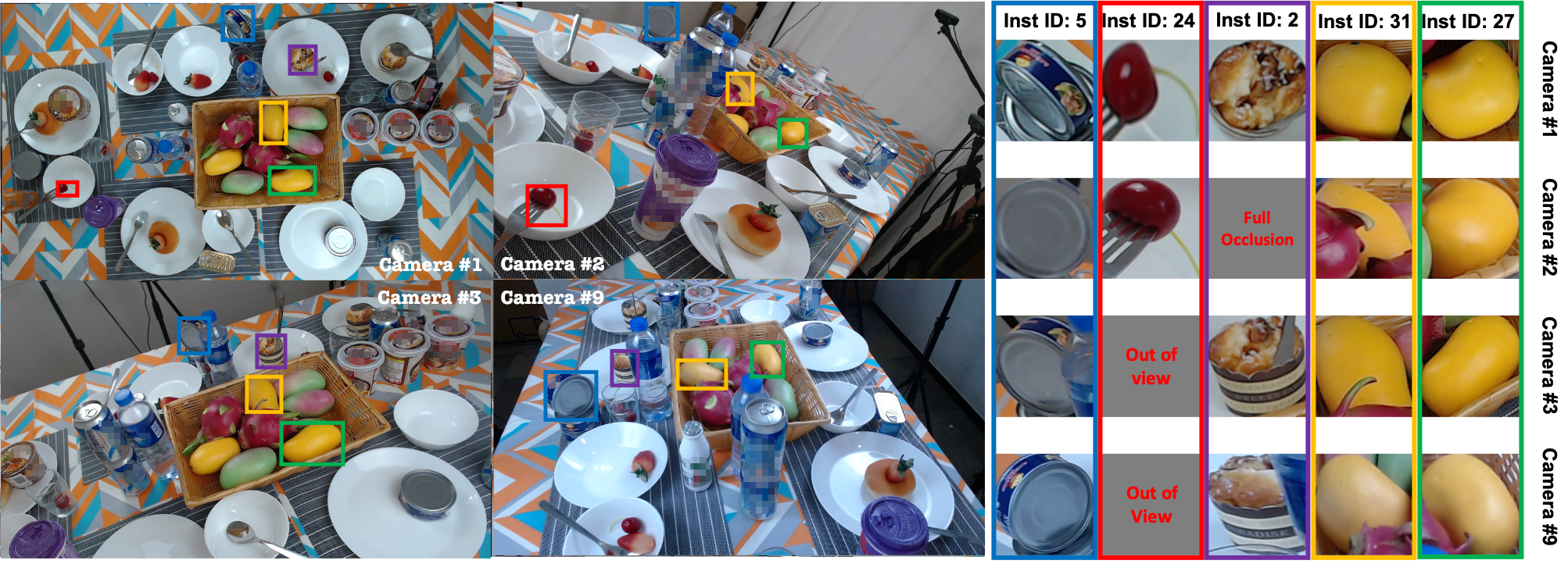}
  \caption{
  MessyTable is a large-scale multi-camera general object dataset with cross-camera association labels. One of the 5,579 scenes is shown in this figure (with only 4 out of 9 cameras depicted). MessyTable is challenging due to severe partial occlusion (instance 2 in camera 9 and instance 31 in camera 2), large camera angle differences, similar or identical-looking instances (instances 31 and 27) and more. Objects are arbitrarily chosen from 120 classes that are commonly found on table. The brand names are pixelated in all illustrations
  }
  \label{fig:messytable_overview}
\end{figure}

\section{Introduction}
\label{sec:intro}

We introduce a new and interesting dataset, {MessyTable}. It contains over 5,000 scenes, each of which captured by nine cameras in one of the 600 configurations of camera poses. Each scene is shot with a random cluttered background and different lighting conditions with about 30 general objects on average. The objects are chosen arbitrarily from 120 classes of possible instances. Figure~\ref{fig:messytable_overview} depicts some scene examples. 
The goal is to associate the different objects in a scene, \ie, finding the right match of the same instance across views.

The seemingly easy task on this dataset is surprisingly challenging.
The relative pose between two cameras can be large, and therefore, an object may appear to be very different when viewed from different angles. 
Objects are heavily occluded while some of them can be elevated by other objects in a cluttered scene. Hence, appearance viewed from different cameras is always partial.
The problem is further complicated with similar-looking or even identical instances.

Solving the aforementioned problems is non-trivial. The geometric constraint is hard to use right away. Multi-view epipolar geometry is ambiguous when a pixel can correspond to all points on the epipolar line in the other view. Homographic projection assumes a reference plane, which is not always available.
To associate an object across camera views, a method needs to distinguish subtle differences between similar-looking objects. Fine-grained recognition is still a challenging problem in computer vision. When a scene contains identical-looking objects, the method is required to search for neighbouring cues in the vicinity of the objects to differentiate them. The neighbouring configuration, however, can be occluded and highly non-rigid with changing relative position due to different camera views.

While the method developed from MessyTable can be applied to some immediate applications such as automatic check-out \cite{wei2019rpc} in supermarkets, \eg, leveraging multiple views to prevent counting error on the merchandise,
the instance association problem found in this dataset is reminiscent of many real-world problems such as person re-identification or object tracking across views. 
Both examples of real-world problems require some sort of association between objects, either through appearances, group configurations, or temporal cues. 
Solving these real-world problems requires one to train a model using domain-specific data. Nonetheless, they still share similar challenges and concerns as to the setting in MessyTable.

MessyTable is not expected to replace the functionalities of domain-specific datasets. It aims to be a general dataset offering fundamental challenges to existing vision algorithms, with the hope of inspiring new problems and encouraging novel solutions. 
In this paper, apart from describing the details of MessyTable, we also present the results of applying some baselines and heuristics to address the instance association problem.
We also show how a deep learning-based method developed from MessyTable can be migrated to other real-world multi-camera domains and achieve good results.


\section{Related Work}

\textbf{Related Problems}. Various computer vision tasks, such as re-identification and tracking, can be viewed as some forms of instance association. Despite differences in the problem settings, they share common challenges as featured in MessyTable, including subtle appearance difference, heavy occlusion and viewpoint variation. We take inspirations from methods in these fields for developing a potential solution for multi-camera instance association.

Re-identification (ReID) aims to associate the query instance (\eg, a person) and the instances in the gallery~\cite{gong2014springer}. Association performance suffers from drastic appearance differences caused by viewpoint variation, and heavy occlusion in crowded scenes. Therefore, the appearance feature alone can be insufficient for satisfactory results. In this regard, instead of distinguishing individual persons in isolation (\eg,\cite{sun2018beyond, zhao2017spindle, zhou2018aware}), an alternative solution proposed by \cite{Zheng2009AssociatingGO} exploits contextual cues: as people often walk in groups in crowded scenes, it associates the same group of people over space and time.

Multi-object Tracking (MOT) is the task to associate instances across sequential frames, leveraging the availability of both appearance features and temporal cues\cite{milan2017online, schulter2017deep, sadeghian2017tracking}. It suffers from ID switches and fragmentation primarily caused by occlusion \cite{luo2014multiple, li2019multiple}. MOT in a multi-camera setting is formally referred to as Multi-Target Multi-Camera Tracking (MTMCT) \cite{ristani2016performance}, which also suffers from viewpoint variation in cross-camera tracklet association \cite{zhang2017multi, gao2018revisiting,hsu2019multi}. In addition, MTMCT with overlapping field of view\cite{xu2016multi, chavdarova2018wildtrack} is similar to MessyTable's  multi-camera setting. Thus, studies conducted on MessyTable might be inspiring for a better cross-camera association performance in MTMCT.

\noindent
\textbf{Related Datasets}. 
Many datasets for ReID and MOT offer prominent challenges \cite{gou2018systematic} that are common in real life. For instance, CUHK03\cite{li2014deepreid}, MSMT17\cite{wei2018person}, and MOT16\cite{milan2016mot16} feature occlusion and viewpoint variation, and many other datasets \cite{ye2020deep, xu2019deep} also feature illumination variations.

There are several multi-camera datasets. Though originally designed for different purposes, they can be used for evaluating instance association. MPII Multi-Kinect (MPII MK) \cite{susanto20123d} is designed for object instance detection and collected on a flat kitchen countertop with nine classes of kitchenwares captured in four fixed views. The dataset features some level of occlusion, but the scenes are relatively simple for evaluating general object association. EPFL Multi-View Multi-Class (EPFL MVMC) \cite{roig2011conditional} contains only people, cars, and buses and is built from video sequences of six static cameras taken at the university campus (with the road, bus stop, and parking slots). WILDTRACK \cite{chavdarova2018wildtrack}, captured with seven static cameras in a public open area, is the latest challenging dataset for multi-view people detection.

Compared to existing datasets, MessyTable aims to offer fundamental challenges that are not limited to specific classes. MessyTable also contains a large number of camera setup configurations for a larger variety of camera poses and an abundance of identical instances that are absent in other datasets.

\section{MessyTable Dataset}
\label{sec:messytable}
\begin{figure}[t]
  \centering
  \includegraphics[width=0.99\textwidth]{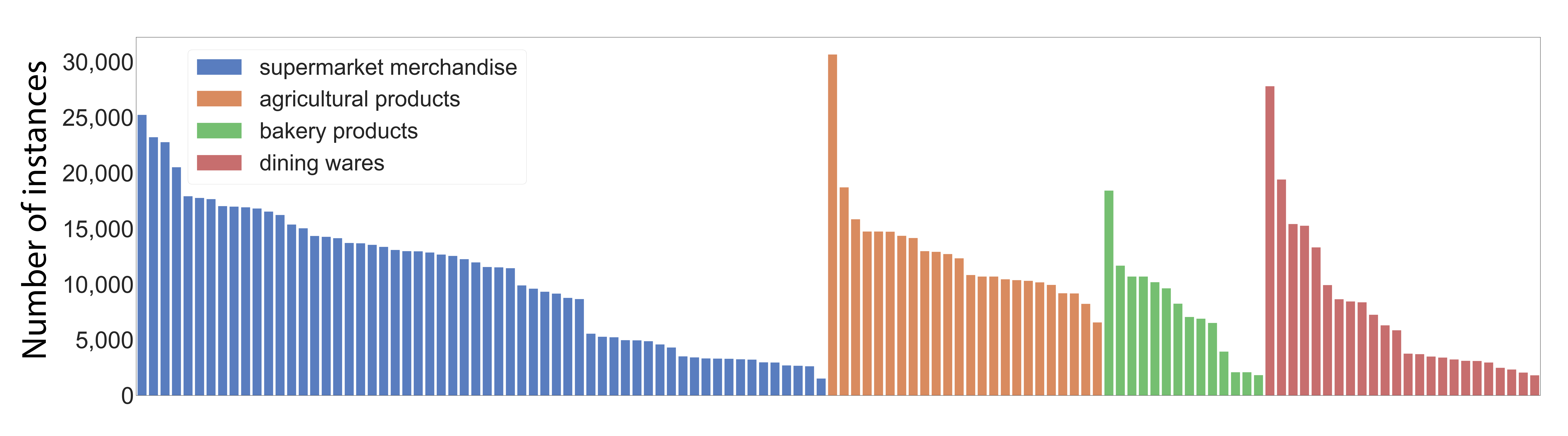}
  \caption{Number of instances per class in MessyTable. MessyTable contains 120 classes: 60 supermarket merchandise, 23 agricultural products, 13 bakery products, and 24 dining wares. A full list of classes is included in the Supplementary Materials}
  \label{fig:instance_frequency}
\end{figure}

MessyTable is a large-scale multi-camera general object dataset designed for instance association tasks. It comprises 120 classes of common on-table objects (Figure \ref{fig:instance_frequency}), encompassing a wide variety of sizes, colors, textures, and materials. Nine cameras are arranged in 567 different setup configurations, giving rise to 20,412 pairs of relative camera poses (Section \ref{sec:variations_in_view_angles}). A total of 5,579 scenes, each containing 6 to 73 randomly selected instances, are divided into three levels of difficulties. Harder scenes have more occlusions, more similar- or identical-looking instances, and proportionally fewer instances in the overlapping areas (Section \ref{sec:variations_in_scenes}). The total 50,211 images in MessyTable are densely annotated with 1,219,240 bounding boxes. Each bounding box has an instance ID for instance association across cameras (Section \ref{sec:data_annotation}). To make MessyTable more representative, varying light conditions and backgrounds are added. Details of the data collection can be found in the Supplementary Materials.

\begin{figure*}[t!]
    \centering
    \includegraphics[width=0.90\linewidth]{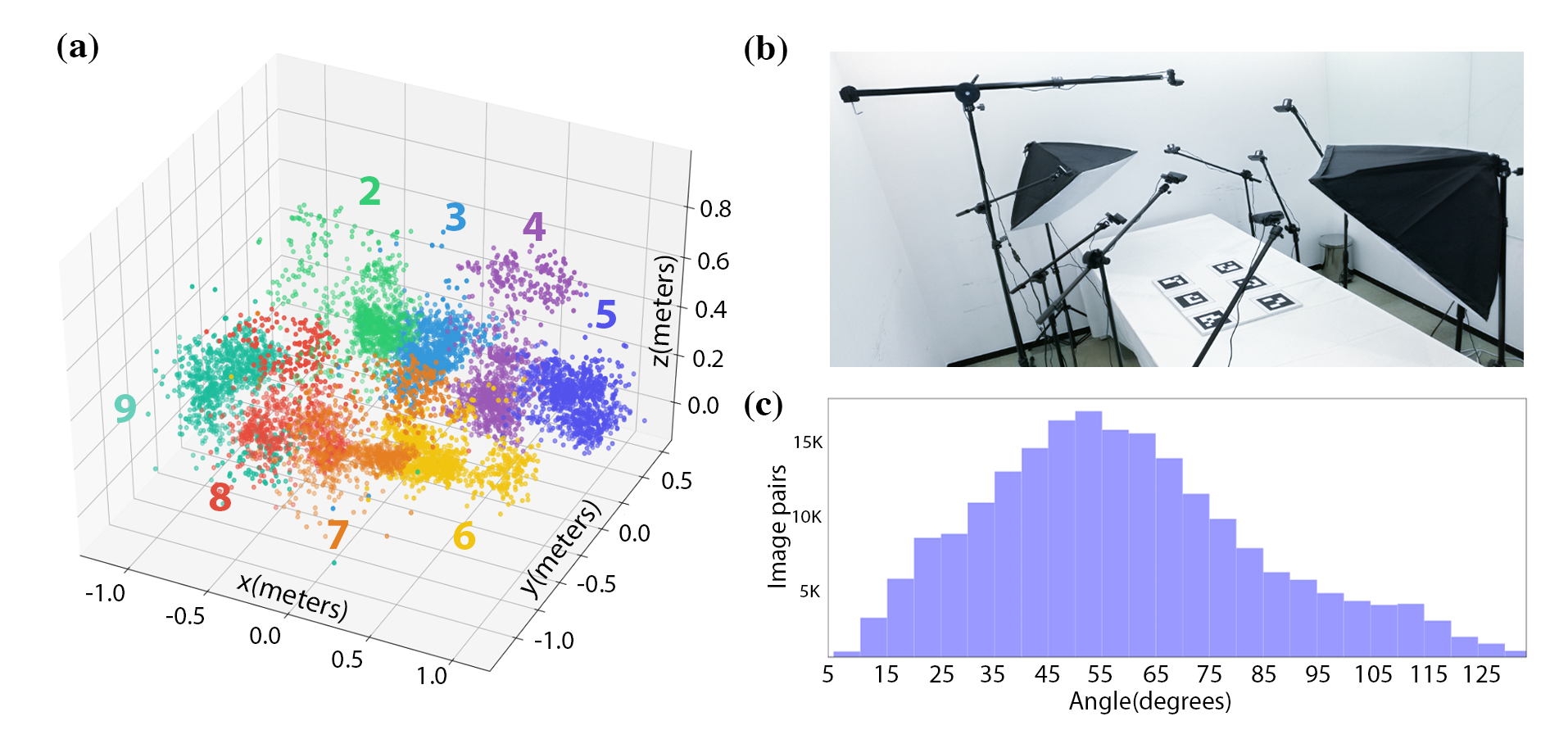}
    \caption{MessyTable has a rich variety of camera poses. (a) The diverse distribution of camera positions covers the entire space. The positions of cameras \#2-9 are projected to the coordinate system of camera \#1, visualized by eight clusters of different colors. Camera \#1 is fixed providing the bird's eye view. (b) The studio for data collection. The checkerboard is used for camera extrinsic calibration (see the Supplementary Materials for details); (c) Distribution of the angle differences of image pairs. The angle is computed between line-of-sights of cameras in each image pair. A significant portion of image pairs have larger angles than that in wide-baseline stereo matching, which is typically capped at 45\degree\cite{caliskan2019learning}} 
    \label{fig:camera_pose}
\end{figure*}

\subsection{Variations in View Angles}
\label{sec:variations_in_view_angles}
For a camera pair with a large angle difference, the same instance may appear to be very different (\eg, instance ID 5 in Figure \ref{fig:messytable_overview}) in the two views. Existing multi-camera datasets typically have their cameras installed on static structures\cite{susanto20123d, roig2011conditional}, even at similar heights\cite{chavdarova2018wildtrack}. This significantly limits the variation as the data essentially collapses to a very limited set of modes. In contrast, MessyTable has not only a high number of cameras but also a large variation in cameras' poses. The camera stands are arbitrarily adjusted between scenes, resulting in an extremely diverse distribution of camera poses and large variance of angle difference between cameras, as shown in Figure \ref{fig:camera_pose}.

\subsection{Variations in Scenes}
\label{sec:variations_in_scenes}

\begin{figure}[t!]
  \centering
  \includegraphics[width=\textwidth]{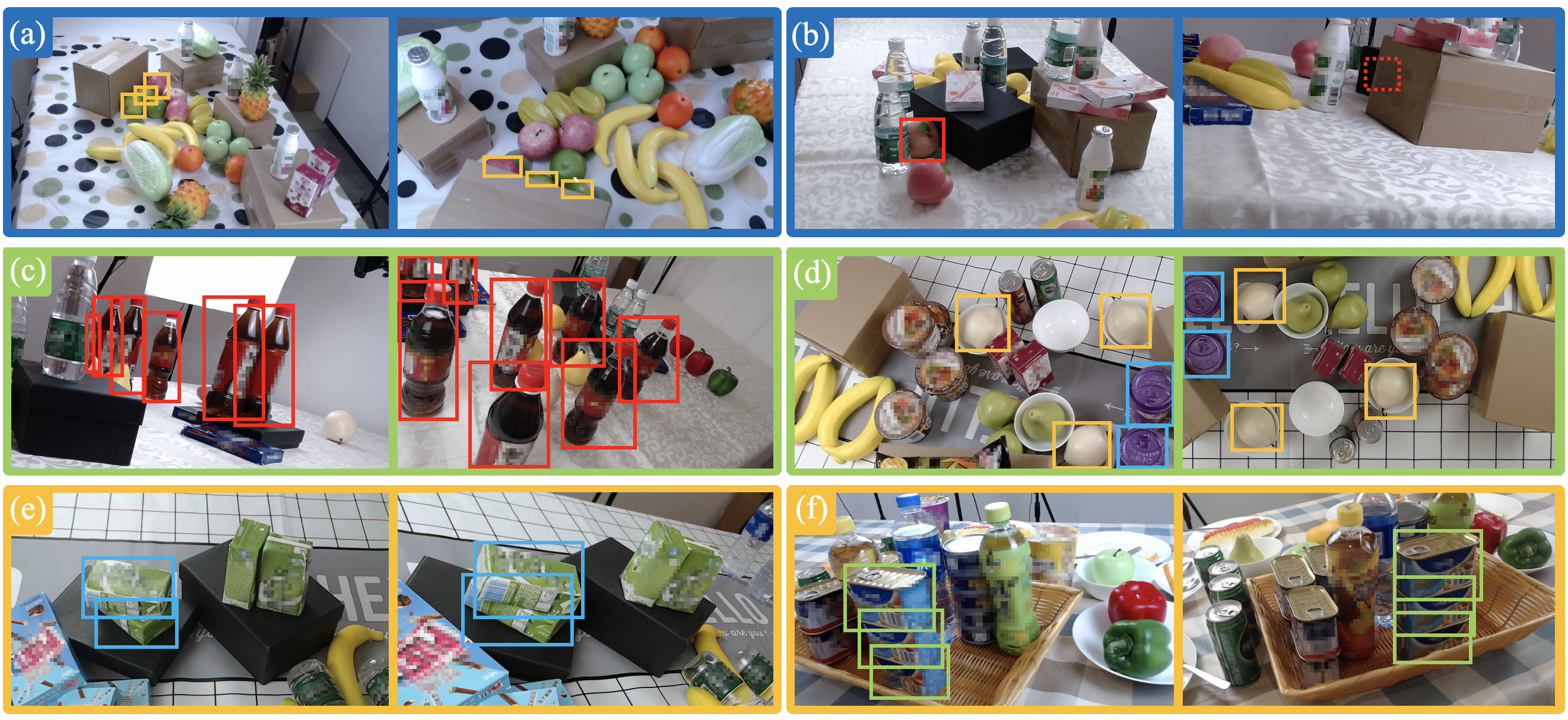}
  \caption{Visualization of some of the design features that impose challenges for association: (a) Partial occlusion. (b) Full occlusion. (c) Similar-looking instances. (d) Identical-looking instances. (e) and (f): Object stacking and placing objects on elevated surfaces}
  \label{fig:visual_design_features}
\end{figure}

\noindent
\textbf{Partial and Full Occlusions}. As shown in Figure \ref{fig:visual_design_features}(a) and (b), partial occlusion results in loss of appearance features \cite{baque2017deep}, making matching more difficult across cameras; full occlusion completely removes the object from one's view despite its existence in the scene. To effectively benchmark algorithms, in addition to dense clutter, artificial obstacles (such as cardboard boxes) are inserted into the scene. 

\noindent
\textbf{Similar- or Identical-looking Instances}. It is common to have similar and identical objects placed in the vicinity as illustrated in Figure \ref{fig:visual_design_features}(c) and (d). Similar or identical instances are challenging for association. MessyTable has multiple duplicates of the same appearance included in each scene, a unique feature that is not present in other datasets such as \cite{susanto20123d}.\

\noindent
\textbf{Variations in Elevation}. Many works simplify the matching problem by assuming all objects are in contact with the same plane \cite{fleuret2007multicamera, chavdarova2017deep, baque2017deep, lopez2018semantic}. However, this assumption often does not hold as the scene gets more complicated. To mimic the most general and realistic scenarios in real life, object instances in MessyTable are allowed to be stacked or placed on elevated surfaces as shown in Figure \ref{fig:visual_design_features}(e) and (f).

\noindent
\textbf{Data Splits}. The dataset is collected with three different complexity levels: Easy, Medium, and Hard, with each subset accounting for 30\%, 50\%, and 20\% of the total scenes. For each complexity level, we randomly partition data equally (1:1:1) into the training, validation, and test sets.

\begin{figure}[t]
\centering
  \includegraphics[width=0.99\textwidth]{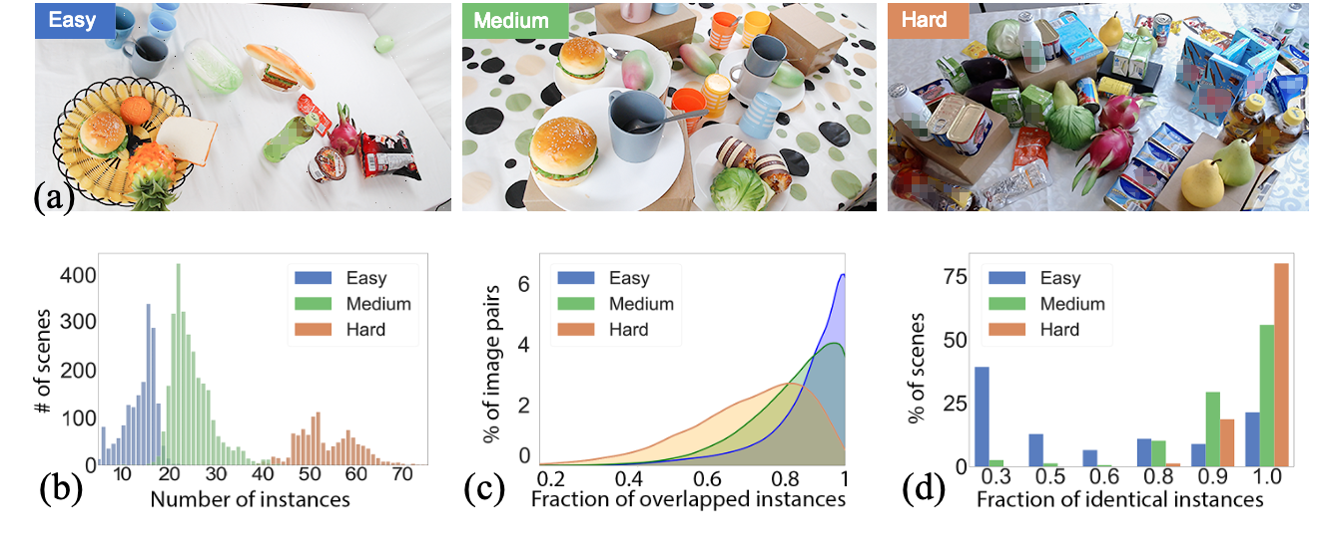}
  \caption{The comparison amongst Easy, Medium and Hard subsets. (a) Example scenes by difficulty: harder scenes have more severe dense clutter and heavy occlusion. (b) Distributions of scenes by the number of instances per scene: harder scenes have more instances. (c) Distribution of image pairs by the extent of instance overlap: the graph peak of harder scenes is shifted leftward, indicating they have less fraction of instances visible in both cameras. (d) Distribution of scenes by the fraction of identical instances in a scene: harder scenes have more identical-looking instances}
  \label{fig:easy_med_hard}
\end{figure}

The camera angle differences are similar among the three levels of complexity. But the number of instances, the fraction of overlapped instances and the fraction of identical instances are significantly different as shown in Figure \ref{fig:easy_med_hard}. Furthermore, as shown in the example scenes, the larger number of instances in a harder scene significantly increases the chance of heavy occlusion. We empirically show that these challenges undermine the association performance of various methods.  

\subsection{Data Annotation}
\label{sec:data_annotation}

We use OpenCV \cite{opencv_library} for calibrating intrinsic and extrinsic parameters. As for the instance association annotation, we gather a team of 40 professional annotators and design a three-stage annotation scheme to obtain reliable annotations in a timely manner. The annotators first annotate bounding boxes to enclose all foreground objects (localization stage), followed by assigning class labels to the bounding boxes (classification stage). In the last stage, we develop an interactive tool for the annotators to group bounding boxes of the same object in all nine views and assign them the same instance ID (association stage). Bounding boxes in different views with the same instance ID are associated (corresponding to the same object instance in the scene) and the ID is unique in each view. For each stage, the annotators are split into two groups in the ratio of 4:1 for annotation and quality inspection.  

It is worth mentioning that the interactive tool has two desirable features to boost efficiency and minimize errors: first, the class labels are used to filter out irrelevant bounding boxes during the association stage; second, the association results correct errors in the classification stage as the disagreement of classification labels from different views triggers reannotation. The details of the data annotation can be found in the Supplementary Materials. In short, MessyTable provides the following annotations:
\begin{itemize}
    \item intrinsic and extrinsic parameters for the cameras;
    \item regular bounding boxes (with class labels) for all 120 foreground objects;
    \item instance ID for each bounding box
\end{itemize}


\section{Baselines}
\label{sec:existing_methods}

In this section, we describe a few popular methods and heuristics that leverage appearance and geometric cues, and a new baseline that additionally exploits contextual information. We adopt a multiple instance association framework, in which pairwise distances between two sets of instances are computed first. After that, the association is formulated as a maximum bipartite matching problem and solved by the Kuhn-Munkres (Hungarian) algorithm. 

\subsection{Appearance Information}

The appearance feature of the instance itself is the most fundamental information for instance association. As the instances are defined by bounding boxes, which are essentially image patches, we find instance association largely resembles the patch matching (local feature matching) problem.

Local feature matching is one of the key steps for low-level multiple camera tasks\cite{Zbontar2015ComputingTS, caliskan2019learning,winder2009picking}. Various hand-crafted feature descriptors, \eg, SIFT \cite{lowe2004distinctive}, have been widely used in this task. We implement a classical matching pipeline including SIFT keypoint description of the instances and K-means clustering for the formation of a visual bag of words (VBoW). The distance between two VBoW representations is computed via chi-square ($\chi^2$). 

The application of deep learning has led to significant progress in patch matching\cite{han2015matchnet, zagoruyko2015learning, csurka2018handcrafted}. The recent works highlight the use of CNN-based discriminative feature extractors such as DeepDesc\cite{simo2015discriminative} to directly output feature vectors, and the distance between two vectors can be computed using L2 distance; MatchNet \cite{han2015matchnet} uses metric networks instead of L2 distance for better performance; DeepCompare\cite{zagoruyko2015learning} proposes to use both multi-resolution patches and metric networks. We use these three works as baselines. 

We also implement a standard triplet network architecture with a feature extractor supervised by the triplet loss during training, which has been proven effective to capture subtle appearance difference in face recognition\cite{schroff2015facenet}. It is referred to as TripletNet in the experiments and L2 distance is used to measure the feature dissimilarity.

\subsection{Surrounding Information}

\begin{figure}[t!]
  \centering
  \includegraphics[width=0.99\textwidth]{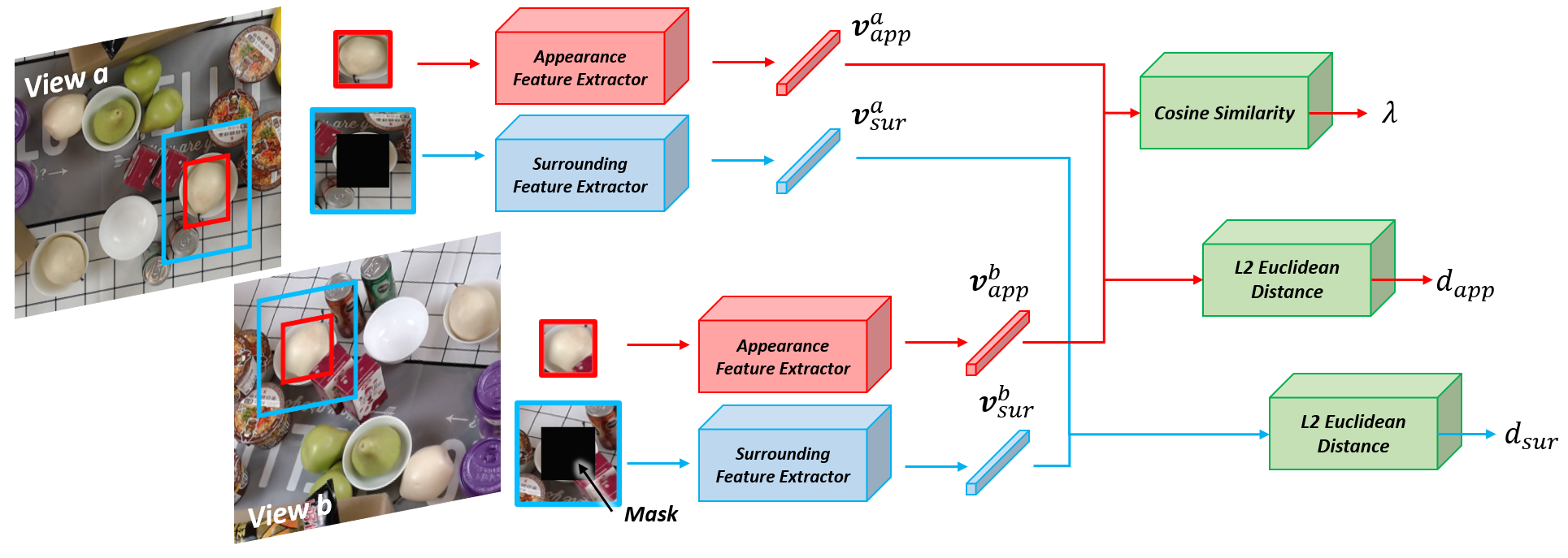}
  \caption{Appearance-Surrounding Network (ASNet) has an appearance branch (red) and a surrounding branch (blue). Each instance and its immediate surrounding are cropped as the input to their respective branches}
  \label{fig:network}
\end{figure}

Inspired by Zheng \etal~\cite{Zheng2009AssociatingGO} in ReID, which addresses occlusion and view variation by associating a group of people instead of the individuals, we propose to look outside of the tight bounding box and involve the neighboring information, hoping to tackle not only occlusion and viewpoint variation, but also the existence of similar-looking instances.

The most intuitive idea is to expand the receptive field by a linear scaling ratio, \ie, cropping an area larger than the actual bounding box. This modification on the network input is referred to in the experiments as \textbf{zoom-out}, and the ratio as \textbf{zoom-out ratio}.

We take further inspiration from the human behavior: one may look at the surroundings for more visual cues \textit{only if} the appearance of the instances themselves are not informative. Hence, we design a simple network (named Appearance-Surrounding Network, Figure \ref{fig:network}), which has two branches for appearance and surrounding feature extraction, fused as follows:
\begin{equation}
d^{ab} = (1 - \lambda) \times D_{l_{2}}(\mathbf{v}_{app}^{a}, \mathbf{v}_{app}^{b}) + \lambda \times D_{l_{2}}(\mathbf{v}_{sur}^{a}, \mathbf{v}_{sur}^{b})
\end{equation}
\begin{equation}
    \lambda = S_{c}(\mathbf{v}_{app}^{a}, \mathbf{v}_{app}^{b})
\end{equation}
\noindent
where $a$ and $b$ are superscripts for two patches, $\mathbf{v}_{app}$ and $\mathbf{v}_{sur}$ are appearance and surrounding feature vectors, respectively. $D_{l_{2}}$ is L2 distance , and $S_{c}$ is cosine similarity. $\lambda$ is the weighting factor to fuse the appearance and surrounding branches. The fusion is designed such that, if the appearance features are similar, the network will place more weight on the surrounding than the appearance. Note that $\lambda$ is jointly optimized in an end-to-end network; it is not a hyperparameter to be set manually.

\subsection{Geometric Methods}
Homographic projection-based methods are very popular and used extensively in past works on Multi-Target Multi-Camera Tracking (MTMCT)\cite{xu2016multi, xu2017cross} and Multi-View People Detection\cite{fleuret2007multicamera, chavdarova2017deep, baque2017deep, lopez2018semantic}. The mid-points of the bottom edges of the bounding boxes \cite{lopez2018semantic} are typically projected to a common coordinate system. The instances can thus be associated based on L2 distance between two sets of projected points. It assumes that all instances are placed on one reference 2D plane (\eg, the ground) and this simplification allows for an unambiguous pixel-to-pixel projection across cameras.

We also make use of epipolar geometry, which does not assume a reference plane. For a pair of bounding boxes in two views, the bounding box center in the first view is used to compute an epipolar line in the second view using the calibrated camera parameters. The distance between the bounding box center in second view and the epipolar line is added to the overall distance between the two bounding boxes. It is a soft constraint, since it does not accept or reject the matches, but penalizes unlikely matches by a large distance.


\section{Experiments}

Unless specified otherwise, we choose ResNet-18 as a light-weight backbone for all models, zoom-out ratio of 2 for models with zoom-out, a mixture of Easy, Medium, and Hard sets are used for training and evaluation.

\subsection{Evaluation Metrics}
\label{sec:eval_metrics}
\noindent
\textbf{AP:} Class-agnostic Average Precision is used to evaluate the algorithm's ability to differentiate positive and negative matches, independent of the choice of the threshold value. All distances are scaled into a range of 0 and 1, and the confidence score is obtained by 1 - x, where x is the scaled distance.

\noindent
\textbf{FPR-95:} False Positive Rate at 95\% recall \cite{han2015matchnet} is commonly used in patch-based matching tasks and is adopted as a supplement to AP. However, it is worth noting that in the patch matching problem, the positive and negative examples are balanced in the evaluation, which is not the case in our task where the negative examples largely outnumber the positive ones.

\noindent
\textbf{IPAA:} We introduce a new metric, Image Pair Association Accuracy (IPAA), that evaluates the image-pair level association results instead of the instance-pair level confidence scores. IPAA is computed as the fraction of image pairs with no less than X\% of the objects associated correctly (written as IPAA-X). In our experiments, we observed that IPAA is more stringent than AP, making it ideal for showing differences between models with reasonably high AP values. Details can be found in the Supplementary Materials.

\subsection{Benchmarking Baselines on MessyTable}
\label{sec:baselines}

\setlength{\tabcolsep}{4pt}
\begin{table}[t!]
\begin{center}
\caption{Baseline performances on MessyTable shows a combination of appearance, surrounding and geometric cues is most effective for instance association. $^{\dagger}$: with metrics learning; $\bigstar$: upgraded backbone for a fair comparison; ZO: zoom-out; ESC: epipolar soft constraint; $^{\ddagger}$: triplet architecture. $\uparrow$: the higher the value, the better; $\downarrow$: the lower the value, the better; the same applies for the rest of the tables and figures}
\label{tab:baselines}
\begin{tabular}{llllll}
\hline\noalign{\smallskip}
Model/Method & AP$\uparrow$ & FPR-95$\downarrow$ & IPAA-100$\uparrow$ & IPAA-90$\uparrow$ & IPAA-80$\uparrow$ \\
\noalign{\smallskip}
\hline
\noalign{\smallskip}
Homography & 0.049  & 0.944 & 0  & 0  & 0 \\
SIFT & 0.063  & 0.866  & 0  & 0  & 0 \\
MatchNet$^{\dagger}$ & 0.193 & 0.458 & 0.010 & 0.012  & 0.033 \\
MatchNet$^{\dagger}\bigstar$ & 0.138 & 0.410 & 0.002  & 0.003  & 0.010 \\
DeepCompare$^{\dagger}$  & 0.202  & 0.412  & 0.023  & 0.025 & 0.063 \\
DeepCompare$^{\dagger}\bigstar$ & 0.129 & 0.402 & 0.005  & 0.005  & 0.01 \\
DeepDesc & 0.090  & 0.906  & 0.011  & 0.011  & 0.018 \\
DeepDesc$\bigstar$  & 0.171  & 0.804  & 0.027  & 0.032  & 0.058 \\
TripletNet$^{\ddagger}$  & 0.467  & 0.206  & 0.168  & 0.220  & 0.376 \\
TripletNet$^{\ddagger}$+ZO  & 0.430  & 0.269 & 0.047  & 0.062  & 0.161 \\
ASNet$^{\ddagger}$  & 0.524  & 0.209  & 0.170  & 0.241  & 0.418 \\
\textbf{ASNet}$^{\ddagger}$\textbf{+ESC} & \textbf{0.577} & \textbf{0.157} & \textbf{0.219}  &
\textbf{0.306}  & \textbf{0.499} \\
\hline
\end{tabular}
\end{center}
\end{table}
\setlength{\tabcolsep}{1.4pt}

In this section, we analyze and provide explanations for the performances of baselines on MessyTable, collated in Table \ref{tab:baselines}. 

Homographic projection performs poorly on MessyTable. The result is not surprising as objects in MessyTable can be placed on different elevated surfaces, violating the 2D reference plane assumption that is critical to accurate projection.

The SIFT-based classical method gives a poor performance as the hand-crafted key points tend to cluster around edges and texture-rich areas, leading to an unbalanced distribution. Hence, texture-less instances have very scarce key points, resulting in ineffective feature representation.

Deep learning-based patch matching SOTAs such as MatchNet \cite{han2015matchnet}, DeepCompare \cite{zagoruyko2015learning}, and DeepDesc \cite{simo2015discriminative} give sub-optimal results as they struggle in distinguishing identical objects, which are abundant in MessyTable. Interestingly, our experiments show that a deeper backbone does not improve performance for MatchNet and DeepCompare, as their performances may be bottlenecked by their simple metric network designs. TripletNet with a triplet architecture outperforms these three models with a Siamese architecture by a clear margin (around a 0.25 increment in AP).

\begin{figure}[t!]
  \centering
  \includegraphics[width=\textwidth]{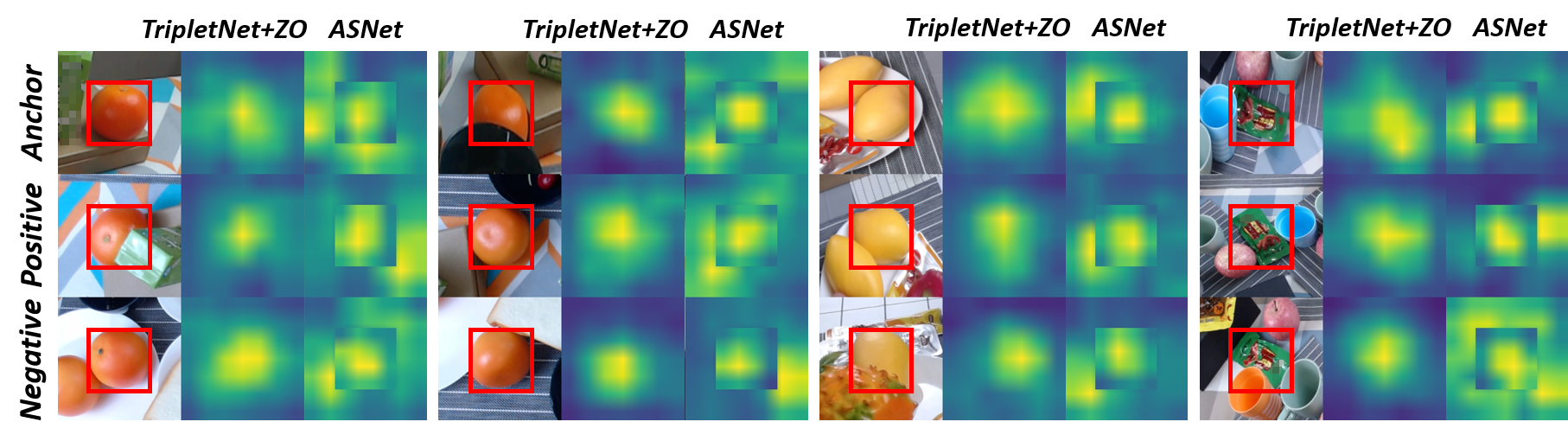}
  \caption{Visualization of feature map activations after conv\_5 of ResNet-18 for TripletNet+ZO (zoom-out) and ASNet. We normalize feature maps to the same scale, then map the values to colors. The higher activation in the surroundings for ASNet indicates that it effectively learns to use surrounding features compared to TripleNet with naive zoom-out}
  \label{fig:featmap_activation}
\end{figure}

We compare TripletNet and ASNet on surrounding information extraction. 
Naive inclusion of surrounding information (TripletNet+ZO) worsens the association results, as a larger receptive field may introduce noises. In contrast, ASNet trains a specialized branch for the surrounding information to extract meaningful features. Figure \ref{fig:featmap_activation} visualizes the feature map activations, showing that ASNet effectively learns to use surrounding information whereas TripletNet+ZO tends to focus on the instance itself. However, we highlight that despite a considerable improvement, the ASNet only achieves a moderate AP of 0.524. This leaves a great potential for improvements. 

We also show that adding soft geometric constraints to ASNet gives further improvement (around 0.05 improvement in AP), indicating that the geometric information is complementary to appearance and surrounding information. However, the performance, especially in terms of the stringent metric IPAA-100, is still unsatisfactory. 

\begin{figure}[t!]
  \centering
  \includegraphics[width=\textwidth]{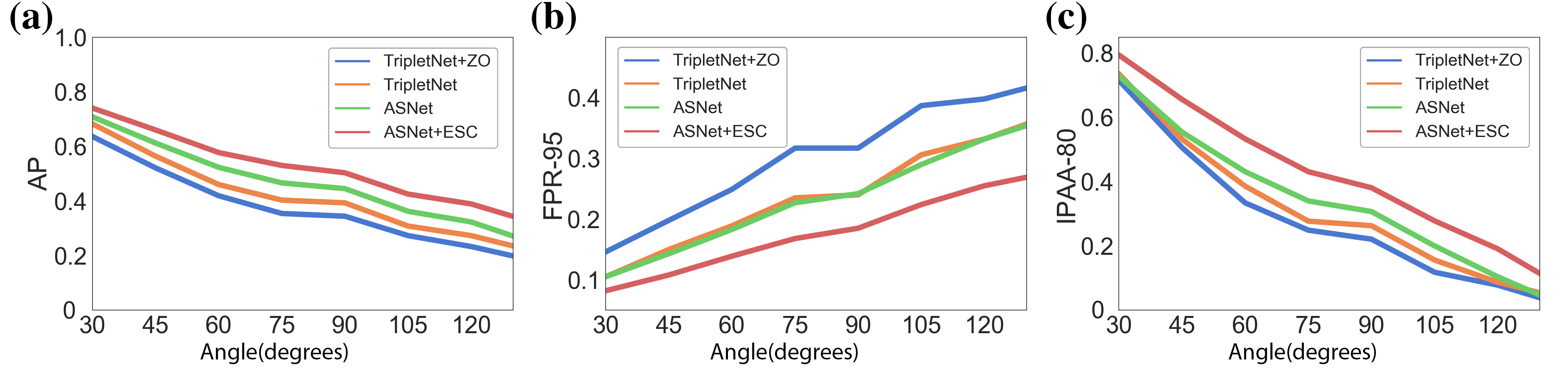}
  \caption{Effects of angle differences. As the angle difference increases, cross-camera association performance deteriorates for all metrics. (a) AP$\uparrow$. (b) FPR-95$\downarrow$. (c) IPAA-80$\uparrow$}
  \label{fig:angle_analysis}
\end{figure}

\subsection{Effects of View Variation and Scene Complexity}
\label{sec:challenges}
We ablate the challenges featured in MessyTable and their effects on instance association. 

We compare the performances of several relatively strong baseline methods at various angle differences in Figure \ref{fig:angle_analysis}. It is observed that the performance by all three metrics deteriorate rapidly with an increase in the angle differences. As shown in Figure \ref{fig:messytable_overview}, large angle difference leads to differences in not only the appearance of an instance itself, but also its relative position within its context. 

\begin{figure}[t!]
  \centering
  \includegraphics[width=0.99\textwidth]{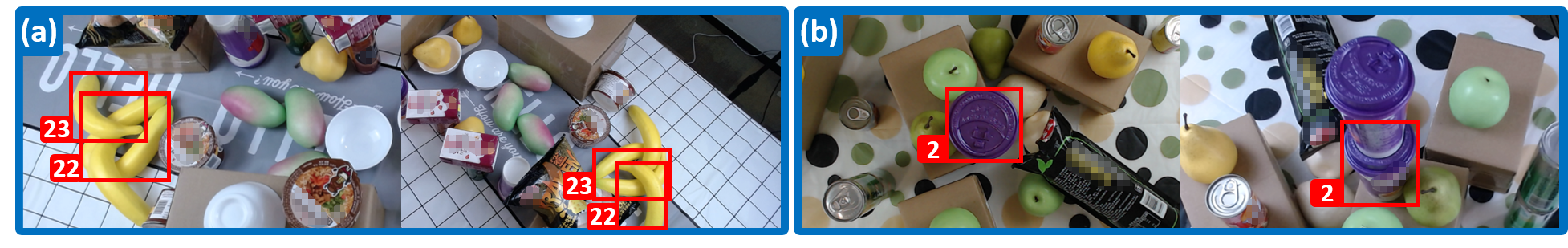}
  \caption{Failure cases: wrong associations are visualized with bounding boxes and instance IDs. Most failure cases occur when identical objects are close to one another such as being (a) placed in a heavy clutter or (b) stacked, leading to indistinguishable surrounding features and similar penalty distances by the soft epipolar constraint. This remains as a challenge worth future research}
  \label{fig:failure_cases}
\end{figure}

In addition, we test the same trained model on Easy, Medium, and Hard test sets. The three test sets have the same distribution of angle differences, but different scene complexity in terms of the number of instances, percentage of identical objects, and the extent of overlapping (Figure \ref{fig:easy_med_hard}). The performance drops significantly in harder scenes, as shown in Table \ref{tab:easy_medium_hard}.
We offer explanations: first, with more instances on the table, harder scenes contain more occlusion, as shown in Figure \ref{fig:easy_med_hard}(a). Second, it is more common to have identical objects closely placed or stacked together, leading to similar surrounding features and geometric distances (Figure \ref{fig:failure_cases}), making such instances indistinguishable. Third, harder scenes have a smaller fraction of instances in the overlapping area, this may lead to more false positive matches between non-overlapped similar or identical objects, which contributes to higher FPR-95 values. 

The above challenges demand a powerful feature extractor that is invariant to viewpoint changes, robust under occlusion, and able to learn the surrounding feature effectively, yet, all baselines have limited capabilities.

\setlength{\tabcolsep}{4pt}
\begin{table}[t!]
\begin{center}
\caption{Results on Easy, Medium, and Hard test sets separately. As the scene get more complicated, the performances of all models worsen. Models are trained on a mixture of Easy, Medium and Hard train sets}
\label{tab:easy_medium_hard}
\begin{tabular}{lllllll}
\hline\noalign{\smallskip}
Subsets & Model & AP$\uparrow$ & FPR-95$\downarrow$ & IPAA-100$\uparrow$ & IPAA-90$\uparrow$ & IPAA-80$\uparrow$ \\
\noalign{\smallskip}
\hline
\noalign{\smallskip}
Easy & TripletNet  & 0.618 & 0.156  & 0.266  & 0.342  & 0.561 \\
& ASNet  & 0.667  & 0.151  & 0.408  & 0.427  & 0.660 \\
& ASNet+ESC  & 0.709  & 0.122  & 0.497  & 0.500  & 0.734 \\
\hline
Medium & TripletNet  & 0.494  & 0.211  & 0.063  & 0.150  & 0.290 \\
& ASNet  & 0.547  & 0.207  & 0.101  & 0.226  & 0.391 \\
& ASNet+ESC  & 0.594  & 0.163  & 0.151  & 0.296  & 0.476 \\
\hline
Hard & TripletNet  & 0.341  & 0.259  & 0.003  & 0.023  & 0.078 \\
 & ASNet  & 0.396  & 0.255  & 0.003  & 0.026  & 0.100 \\
 & ASNet+ESC  & 0.457  & 0.185 & 0.007  & 0.048   & 0.155 \\
\hline
\end{tabular}
\end{center}
\end{table}
\setlength{\tabcolsep}{1.4pt}

\subsection{MessyTable as a Benchmark and a Training Source}
\label{sec:other_datasets}

We further validate the usefulness of MessyTable by conducting experiments on three public multi-camera datasets (Table \ref{tab:other_datasets}), which gives the following insights:

First, methods that saturate MPII MK and EPFL MVMC are far from saturating MessyTable (Table \ref{tab:baselines}). Note that both datasets have a limited number of classes and instances. Hence, this result highlights the need for MessyTable, a more realistic and challenging dataset for research of instance association.

Second, it is observed that algorithms show consistent trends on MessyTable and other datasets, that is, an algorithm that performs better on MessyTable also performs better on all other datasets. This shows MessyTable can serve as a highly indicative benchmark for multi-camera instance association. 

Third, models pretrained on MessyTable consistently perform better than those pretrained on ImageNet, showing MessyTable is a better training source for instance association tasks. Note that EPFL MVMC has three classes (people, cars, and buses) and WILDTRACK is a challenging people dataset. It shows that a model trained on the general objects in MessyTable, learns feature extraction that is readily transferable across domains without feature engineering. 

\setlength{\tabcolsep}{4pt}
\begin{table}[t!]
\begin{center}
\caption{Experiments on three other multi-camera datasets. MessyTable is highly indicative as algorithms benchmarked on MessyTable show consistent trends on all other datasets. Moreover, models pretrained on MessyTable consistently give better performances, showing MessyTable's potential as a training source}
\label{tab:other_datasets}
\begin{tabular}{llllll}
\hline\noalign{\smallskip}
Dataset & Model & Pretraining & AP$\uparrow$ & FPR-95$\downarrow$ & IPAA-100$\uparrow$  \\
\noalign{\smallskip}
\hline
\noalign{\smallskip}
MPII MK & TripletNet & ImageNet & 0.847  & 0.196 & 0.333\\
 & ASNet & ImageNet & 0.881 & 0.167 & 0.696 \\
 & ASNet & MessyTable & 0.905 & 0.119 & 0.765\\
\hline
EPFL MVMC & TripletNet & ImageNet & 0.921 & 0.056 & 0.529 \\
 & ASNet & ImageNet & 0.950 & 0.038 & 0.559  \\
 & ASNet & MessyTable & 0.969 & 0.031 & 0.575 \\
\hline
WILDTRACK & TripletNet & ImageNet & 0.616 & 0.207 & 0.095 \\
 & ASNet & ImageNet & 0.718 & 0.094 & 0.304 \\
 & ASNet & MessyTable & 0.734 & 0.083 & 0.321 \\
\hline
\end{tabular}
\end{center}
\end{table}
\setlength{\tabcolsep}{1.4pt}


\section{Discussion}
In this work, we have presented MessyTable, a large-scale multi-camera general object dataset for instance association. MessyTable features the prominent challenges for instance association such as appearance inconsistency due to view angle differences, partial and full occlusion, similar and identical-looking objects, difference in elevation, and limited usefulness of geometric constraints. We show in the experiments that it is useful in two more ways. First, MessyTable is a highly indicative benchmark for instance association algorithms. Second, it can be used as a training source for domain-specific instance association tasks.

By benchmarking baselines on MessyTable, we obtain important insights for
instance association: appearance feature is insufficient especially in the presence
of identical objects; our proposed simple baseline, ASNet, incorporates the surrounding information into association and effectively improves the association
performance. In addition, we show that epipolar geometry as a soft constraint
is complementary to ASNet.

Although the combined use of appearance features, context information, and geometric cues achieves reasonably good performance, ASNet is still inadequate to tackle all challenges. Therefore, we ask three important questions: (1) how to extract stronger appearance, neighbouring and geometric cues? (2) is there a smarter way to fuse these cues? (3) is there more information that we can leverage to tackle instance association?

The experiment results on MessyTable set many directions worth exploring. First, increasing view angle difference leads to a sharp deterioration of instance association performance of all baselines, highlighting the need for research on methods that capture view-invariant features and non-rigid contextual information. Second, methods give poorer performances as the scenes get more complicated; failure cases show that identical instances placed close to each other are extremely difficult to address despite that the strongest baseline already leverages appearance, surrounding and geometric cues. Hence, more in-depth object relationship reasoning may be helpful to distinguish such instances. 

\noindent
\textbf{Acknowledgements}. This research was supported by SenseTime-NTU Collaboration Project, Singapore MOE AcRF Tier 1 (2018-T1-002-056), NTU SUG, and NTU NAP.
\setcounter{section}{0}
\renewcommand\thesection{\Alph{section}}
\section*{\fontsize{15}{15}\selectfont Appendix}
\section{Content Summary}
In the supplementary materials, we provide additonal details on:
\begin{itemize}
    \item data collection procedure;
    \item data annotation procedure;
    \item full list of the 120 classes of objects;
    \item example scenes of three difficulty levels: Easy, Medium, and Hard;
    \item statistics of MessyTable and the three datasets evaluated in Section 5.4;
    \item framework;
    \item proposed metric IPAA;
    \item baselines
\end{itemize}
\section{Additional Details on Data Collection}
\label{sec:dataset:data_collection}
We gather a team of 10 people for data collection, we refer to them as data collectors. We define the term ``setup" and ``scene" as follows: a setup is an arrangement of nine cameras. The camera poses are randomly set for a setup and are reset for subsequent setups. A scene is an arrangement of all objects on the table: a random set of objects are being placed on the table. These objects are then cleared from the table and replaced with a new random set of objects for subsequent scenes. With each setup, each camera captures one photo for each scene; a total of 10 scenes are collected for each setup.

\subsection{Setup}
\noindent
\textbf{Camera Poses and Extrinsic Calibration}
For each setup, cameras poses, except camera \#1 that provides a bird’s eye view of the scene, are varied. Certain camera poses are deliberately arranged to be very near the table surface, to collect images of an incomplete scene. A calibration board, with six large ArUco\cite{romero2018speeded,garrido2016generation} markers are then placed on the table, at a position that is visible to all cameras. The detected marker corners are used to compute the transformation matrix from the board frame to the camera frame by solving the the perspective-n-points problem \cite{opencv_library}.

\noindent
\textbf{Lighting Conditions} Variations in lighting often severely affect the performances of visual algorithms. Data augmentation \cite{shorten2019survey} and artificially generated shadows \cite{wei2019rpc} can be unrealistic. Hence, we combine fixed light sources with mobile studio lighting kits to add lighting variations to the dataset such as different light directions and intensity, shadows, and reflective materials. The lighting is adjusted for every setup.

\subsection{Scene}
For object placements, we only provide vague instructions to the data collectors about the approximate numbers of objects to be used for Easy, Medium, and Hard scenes respectively; the data collectors make their own decisions at choosing a set of objects and the pattern to place the objects on the table. Hence, we ensure that the object placements resemble the in-the-wild arrangements as much as possible.

\noindent
For backgrounds, we include baskets and cardboard boxes during data capturing. They serve various purposes, including as occlusion, as platforms for other objects, etc. We also have coasters, placemats, and tablecloths underneath each scene which come in different sizes, patterns, colors, and textures, and are commonly found in natural scenes.

\section{Additional Details on Data Annotation}
\label{sec:dataset:data_annotation}

\begin{figure}[t!]
  \centering
  \includegraphics[width=0.99\textwidth]{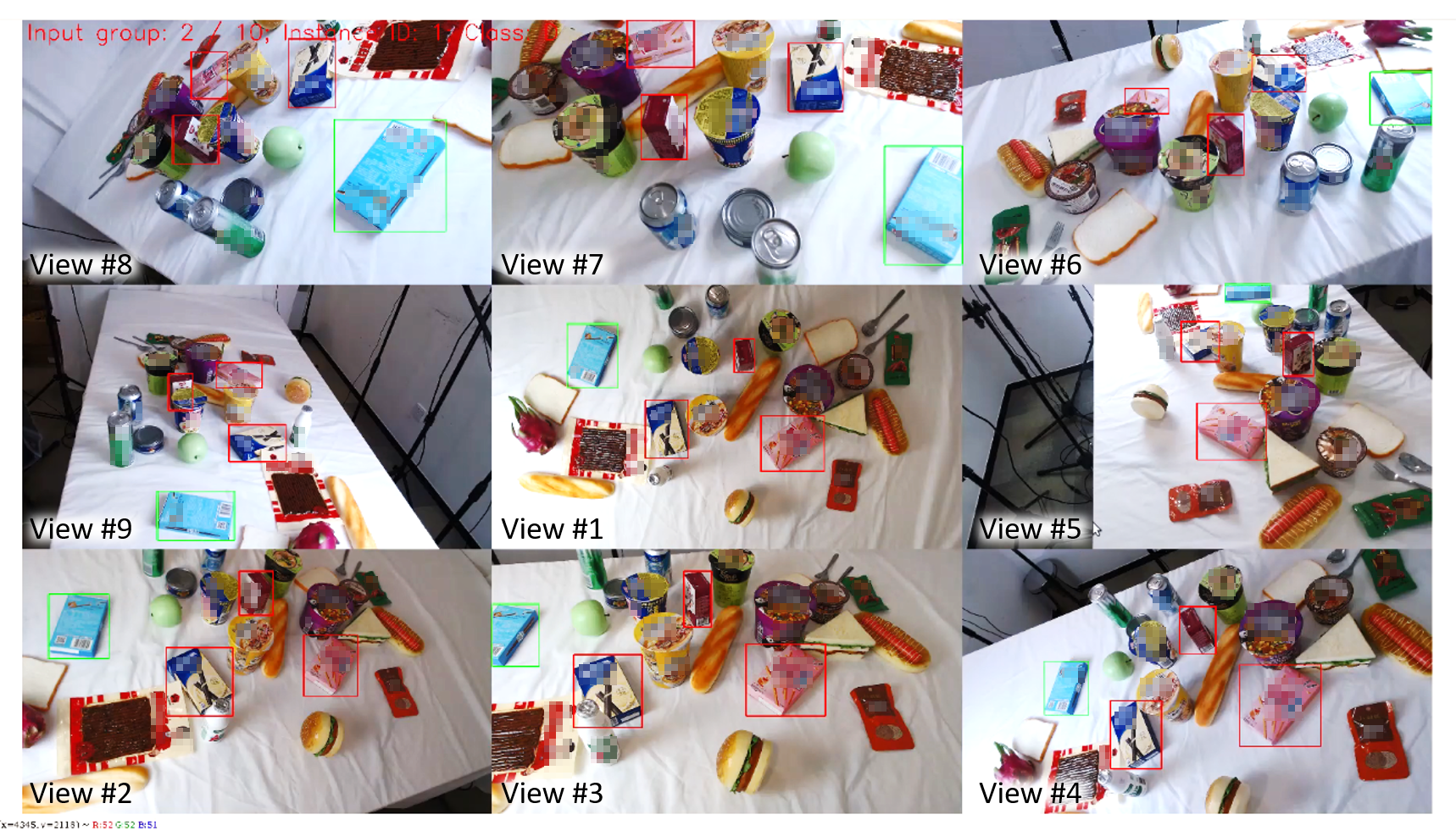}
  \caption{The user interface of the interactive tool. The views are arranged according to the actual camera locations. The green bounding boxes are currently selected to be assigned the same instance ID. The red bounding boxes have similar class labels. The rest of bounding boxes are not displayed. The brand names are pixelated in all illustrations.}
  \label{fig:tool}
\end{figure}

The interactive tool we design for the association stage is shown in Figure \ref{fig:tool}. By selecting bounding boxes, these bounding boxes are assigned the same instance ID. The tool is designed with the following features to increase efficiency and to minimize errors:

\noindent
\textbf{Irrelevant Bounding Box Filtering}
Once a bounding box is selected (by clicking on it) in any view, only the bounding boxes of the same class or similar classes will remain displayed in other views. It is worth noting that we choose to keep similar classes, in addition to the same class, because the labels from the classification stage can be erroneous (a object is wrongly annotated with a similar class to the true class). Classes are considered to be similar based on their categories (the grouping is listed in Table \ref{tab:grouping}). 

\noindent
\textbf{Classification Annotation Verification}
The tool checks if the bounding boxes with the same instance ID have the same class labels. It notifies annotators if any disagreement is detected, and performs automatic correction based on majority voting of the class label amongst nine views, each annotated independently in the classification stage.

\setlength{\tabcolsep}{4pt}
\begin{table}[h!]
\begin{center}
\caption{Grouping of classes used in the association annotation stage to accelerate the annotation by filtering out irrelevant bounding boxes}
\label{tab:grouping}
\begin{tabular}{lll}
\hline\noalign{\smallskip}
Group & Class & Description \\
\noalign{\smallskip}
\hline
\noalign{\smallskip}
A & 1-10 & bottled drinks \\ 
B & 11-19 & cupped food \\
C & 20-30 & canned food \\
D & 31-41 & boxed food \\
E & 42-50 & vacuum-packed food \\
F & 51-60 & puffed food \\
G & 61-77 & fruits \\
H & 78-83 & vegetables \\
I & 84-96 & staples \\
J & 97-100 & utensils \\
K & 101-107 & bowls \& plates \\
L & 108-115 & cups \\
M & 116-120 & drink glasses \\
\hline
\end{tabular}
\end{center}
\end{table}
\setlength{\tabcolsep}{1.4pt}

\clearpage
\section{Full List of 120 Object Classes}
\label{sec:full_list}

\begin{figure}[h!]
  \centering
  \includegraphics[width=0.98\textwidth]{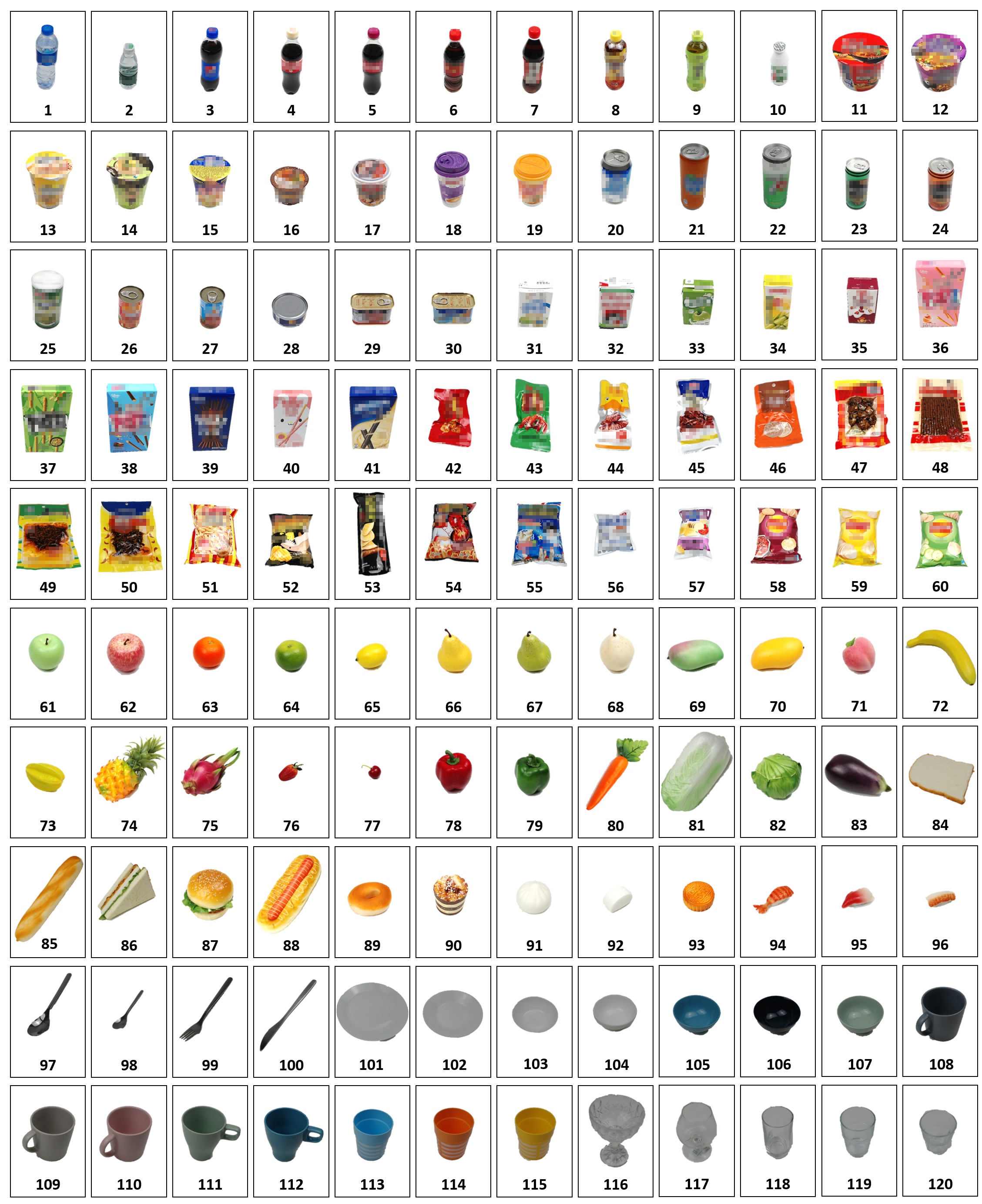}
  \caption{The full list of the 120 classes of objects. The objects are commonly found on a table in real life. They have a wide variety of sizes, colors, textures, and materials. Supermarket merchandise: 1-60; agricultural products: 61-83; bakery products: 84-96; dining wares: 97-120. Note that highly realistic food models are used for class 61-96 as the actual food is perishable, making it not suitable for data collection which spans over a few months}
  \label{fig:full_list}
\end{figure}

\clearpage
\section{Example Scenes}
\label{sec:full_scenes}

\begin{figure}[h!]
  \centering
  \includegraphics[width=0.75\textwidth]{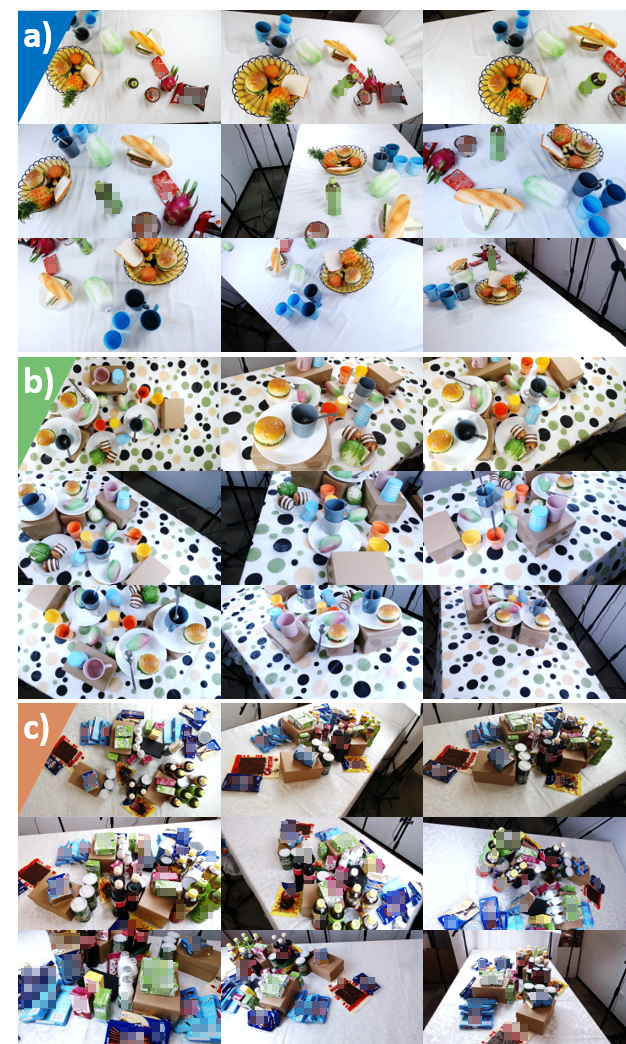}
  \caption{Example scenes in all nine views. (a) An Easy scene with 19 objects. (b)A Medium scene with 27 objects. (c)A Hard scene with 56 objects. Harder scenes have more object instances, more severe occlusion, and more similar/identical objects. Only part of the scene is visible in some camera poses}
  \label{fig:full_scenes}
\end{figure}

\clearpage
\section{Additional Statistics of MessyTable and Other Datasets}
\label{sec:stats}

Table \ref{tab:datasets_comparisons} shows the additional statistics of MessTable and the three datasets that were evaluated in Section 5.4.

\setlength{\tabcolsep}{4pt}
\begin{table}
\begin{center}

\caption{Comparison with other multi-camera datasets. MessyTable is the largest in all aspects below. 
}
\label{tab:datasets_comparisons}
\begin{tabular}{lllllllll}
\hline\noalign{\smallskip}
Datasets & Classes  & Cameras & Setups & Scenes & Images & BBoxes & Instances \\
\noalign{\smallskip}
\hline
\noalign{\smallskip}
MPII MK            & 9   & 4 & 2 & 33   & 132  & 1,051 & 6-10\\
EPFL MVMC          & 3   & 6 & 1 & 240 & 1,440 & 4,081 & 5-9\\
WILDTRACK          & 1   & 7 & 1 & 400 & 2,800 & 42,707 & 13-40 \\
\textbf{MessyTable} & \textbf{120} & \textbf{9} & \textbf{567} & \textbf{5,579}  & \textbf{50,211} & \textbf{1,219,240} & \textbf{6-73}\\
\hline
\end{tabular}
\end{center}
\end{table}
\setlength{\tabcolsep}{1.4pt}

\section{Additional Details on the Framework}
\label{sec:framework}

\begin{figure}
  \centering
  \includegraphics[width=\textwidth]{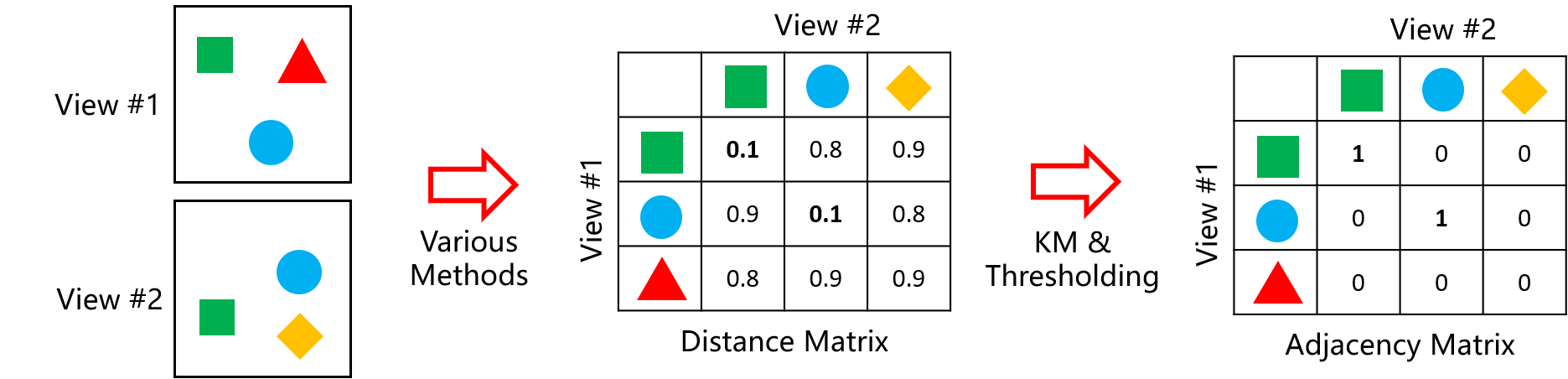}
  \caption{The general framework for instance association in a multi-camera setting. In this example, the red triangle is only visible in View \#1 and the yellow diamond is only visible in View \#2. All methods we explain in the main paper essentially compute pair-wise distances between instances. KM stands for Kuhn-Munkres algorithm, which globally optimizes the matches such that the total loss (the sum of distances of matched pairs) is the minimum. An additional thresholding step further rejects matches with large distances}
  \label{fig:framework}
\end{figure}

As shown in Figure \ref{fig:framework}, all baselines discussed in the main paper are essentially different ways to compute the pair-wise distances. Homographic projection uses the pixel distance between two sets of projected points; SIFT uses the chi-square distance between two visual bag of words representations; MatchNet and DeepCompare use metric networks to compute the similarity between extracted feature vectors; DeepDesc, TripletNet, and ASNet use L2 distance; Epipolar soft constraint uses pixel distance between a bounding box center point and an epipolar line.


\section{Additional Details on the Proposed Metric: \\ Image-pair Association Accuracy (IPAA)}
\label{sec:ipaa}

\begin{figure}[t!]
  \centering
  \includegraphics[width=0.99\textwidth]{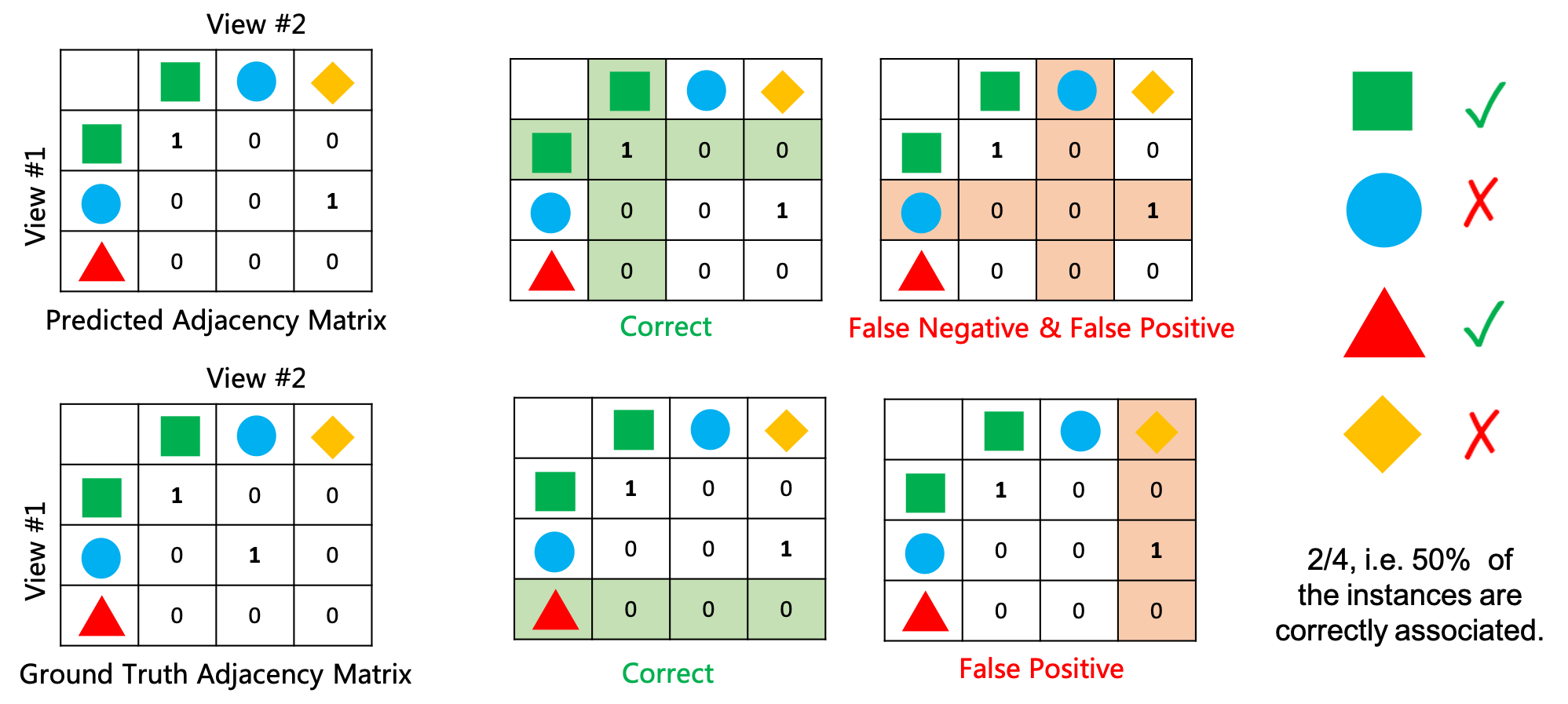}
  \caption{Computation of the percentage of correctly associated objects in an image pair. The predicted adjacency matrix (Section \ref{sec:framework}) is compared against the ground truth for each object present in either of the two images. IPAA-X is the fraction of image pairs that have no less than X\% of objects associated correctly}
  \label{fig:ipaa}
\end{figure}

The motivation for IPAA is to gauge performance at the image-pair level whereas AP and FPR-95 gauge performance at the instance-pair level: AP and FPR-95 evaluate the matching score (confidence score) of each instance pair against its ground truths (0 or 1), but do not directly provide insights of the matching quality of an image pair, which contain many instance pairs. In contrast, IPAA is computed as the fraction of image pairs with no less than X\% of the objects associated correctly (written as IPAA-X). The computation of the percentage of correctly associated objects for each image pair is shown in Figure \ref{fig:ipaa}.

\section{Additional Details on Baselines}

This section provides more details on baselines. These details are excluded in the main paper due to space constraint, but they offer important insights on the instance association problem.

\begin{figure}[t!]
  \centering
  \includegraphics[width=0.95\textwidth]{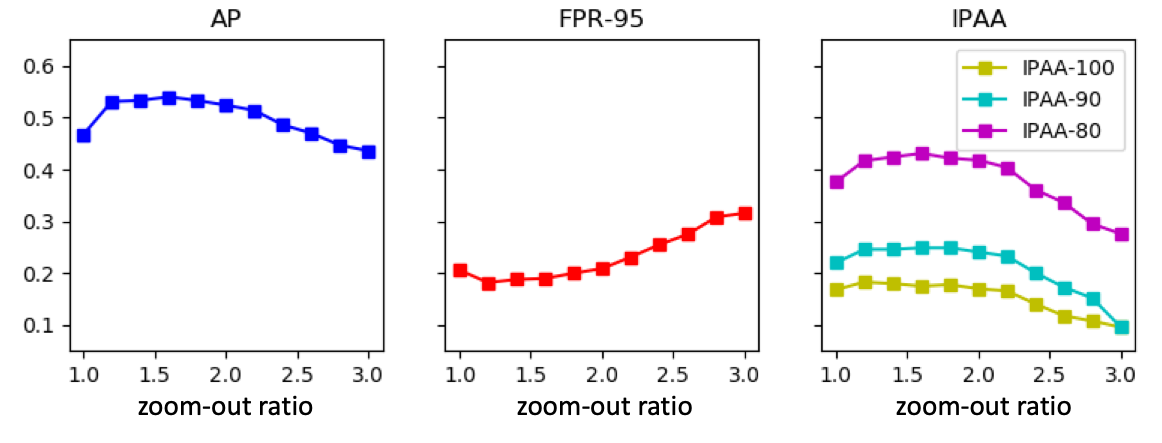}
  \caption{Performance of ASNet is not sensitive to the value of zoom-out ratio in the range [1.2, 2.2], after which it drops rapidly} 
  \label{fig:zoom_out_raio}
\end{figure}

\subsection{Additional Results on Zoom-out Ratio}

By including surrounding information, the key hyperparameter for our baseline ASNet is the zoom-out ratio. We also conduct experiments on different zoom-out ratios. It shows that including surrounding information significantly improves the association performance (compared to that when zoom-out ratio = 1). We simply choose the zoom-out ratio to be 2 as the performance is not sensitive to the value of zoom-out ratio in the range [1.2, 2.4]. 
However, as the zoom-out ratio increases beyond 2.4, the performance starts to decline. We argue that even though a larger zoom-out ratio could include more surrounding area, the model is unable to extract an effective embedding for the surrounding features. This can be a direction for future research.  

\setlength{\tabcolsep}{4pt}
\begin{table}[t!]
\begin{center}
\caption{Instance association performance of ASNet using detected bounding boxes. The instance association performance suffers from imperfect bounding boxes generated by detectors compared to ground truth bounding boxes. The performance deteriorates as the detectors become weaker}
\label{tab:det}
\begin{tabular}{lllll}
\hline\noalign{\smallskip}
Detector & Detection mAP$\uparrow$ & IPAA-100$\uparrow$ & IPAA-90$\uparrow$ & IPAA-80$\uparrow$\\
\noalign{\smallskip}
\hline
\noalign{\smallskip}
\textbf{GT Bounding Box} & \textbf{1.0} & \textbf{0.170}  & \textbf{0.241}  & \textbf{0.418} \\
Cascade Faster R-CNN r101 & 0.797 & 0.153 & 0.212 & 0.388 \\
Cascade Faster R-CNN r50 & 0.772 & 0.141 & 0.198 & 0.366\\
Faster R-CNN r101 & 0.756 & 0.120 & 0.165 & 0.326 \\
Faster R-CNN r50 & 0.722 & 0.097 & 0.135 & 0.283 \\

\hline
\end{tabular}
\end{center}
\end{table}
\setlength{\tabcolsep}{1.4pt}

\subsection{More Details on Using Bounding Boxes from Detectors}

We also evaluate our trained ASNet model on the test set where the bounding boxes are generated by detectors, instead of the ground truth bounding boxes. These detected bounding boxes suffer from false positive (false detection), false negative (missed detection), and imperfect localization and dimension.

It is worth noting that the detected bounding boxes undergo post-processing to obtain instance IDs from the ground truth. For a given image, bipartite matching is performed between the detected bounding boxes and the ground truth bounding boxes based on pair-wise IoUs. The matched detected bounding boxes are assigned the instance IDs of the ground truth bounding boxes, whereas the unmatched detected bounding boxes are assigned unique instance IDs.

The results are collated in Table \ref{tab:det}. Instance association itself is challenging, let alone combining it with a detection stage. The weaker the detection model used as the upstream, the worse the association performance gets. We point out that joint optimization of the detection and the association stage can be a direction for future research.

\begin{figure}[t!]
  \centering
  \includegraphics[width=0.99\textwidth]{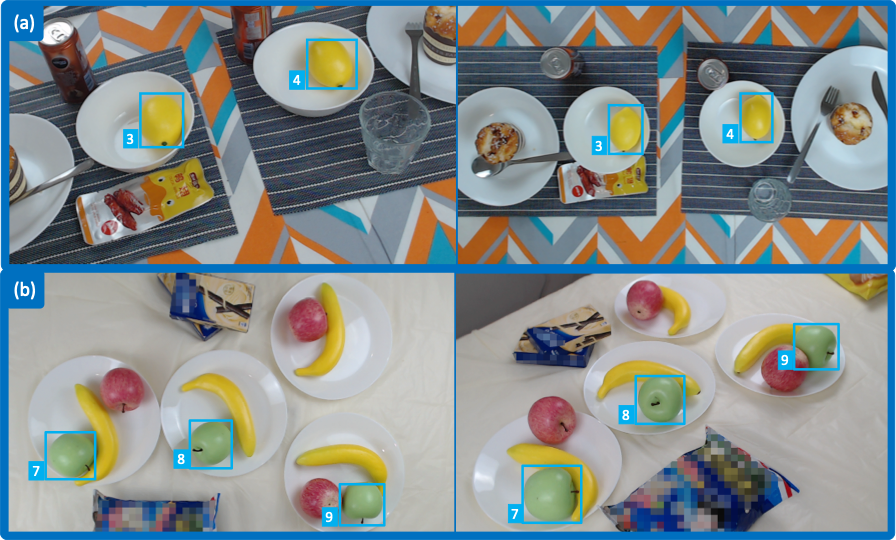}
  \caption{Visualization of cases where both appearance features and surrounding features combined are insufficient for instance association. In this regard, the soft epipolar constraint is necessary as it assigns the geometrically infeasible pair (\ie, false pair) a larger distance}
  \label{fig:geometric_needed_cases}

\end{figure}

\begin{figure}
  \centering
  \includegraphics[width=0.99\textwidth]{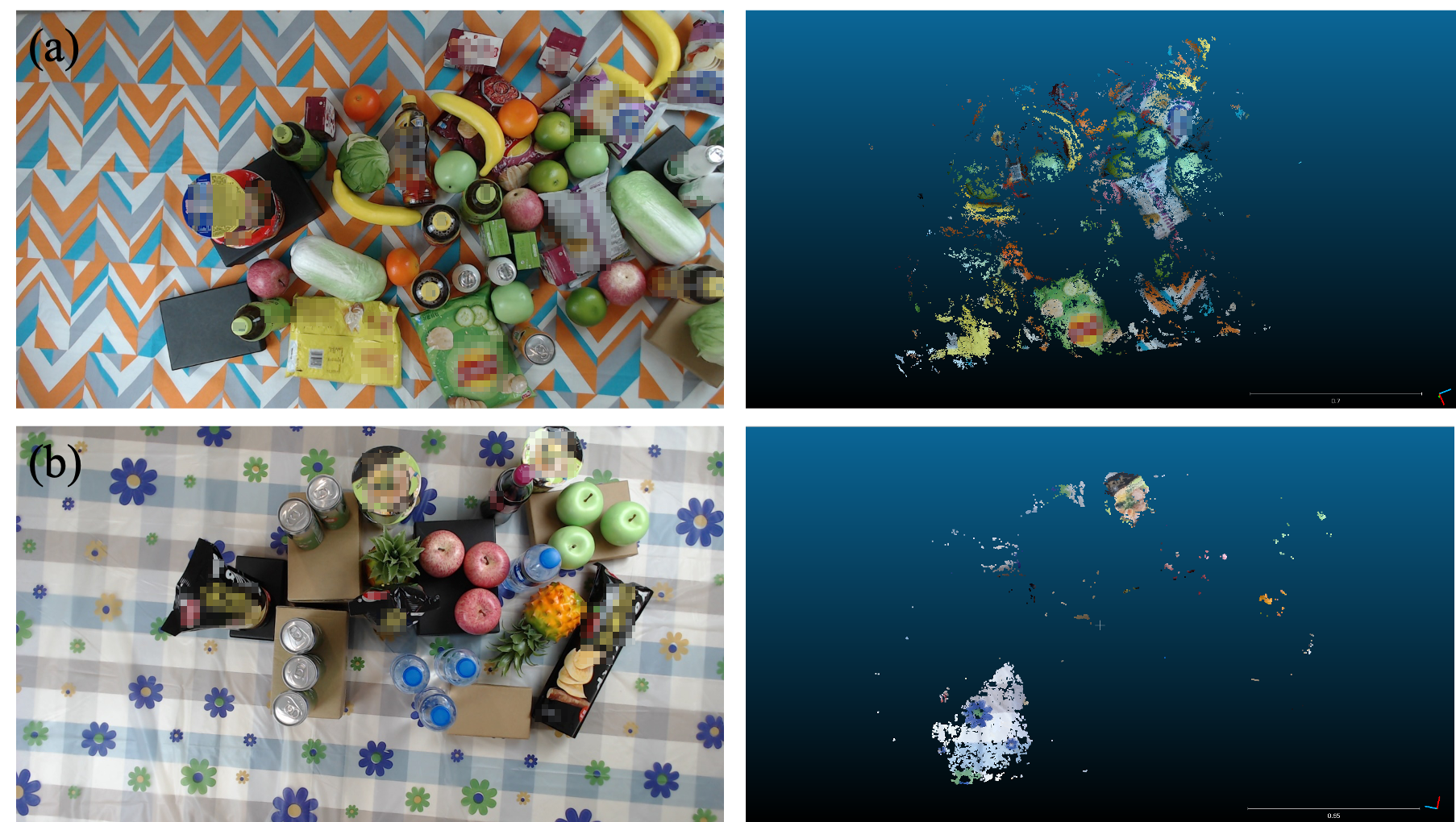}
  \caption{(a) and (b) are two examples of 3D reconstruction results: view \#1 of the scene is placed on the left and the construction result on the right. SfM is performed by Theia\cite{theia-manual} and multi-view stereo is performed by OpenMVS\cite{cernea2015openmvs}}
  \label{fig:sfm_failures}
\end{figure} 

\begin{figure}
  \centering
  \includegraphics[width=0.8\textwidth]{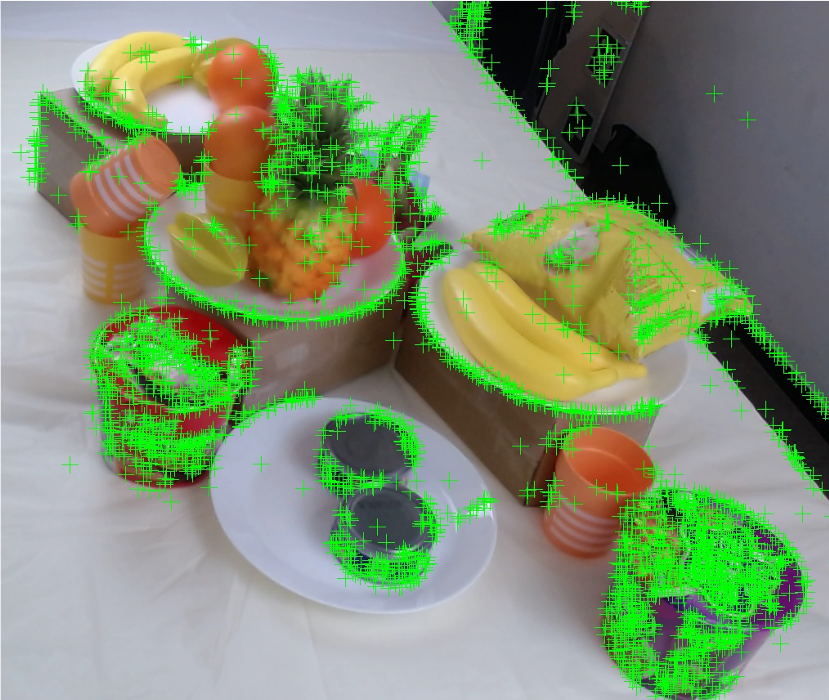}
  \caption{SIFT keypoints have an imbalanced distribution among instances. There are instances with few keypoints, \eg, the yellow cup in the image}
  \label{fig:sift}
\end{figure}

\subsection{Additional Visualization of Scenes Where Geometric Cues Are Necessary}

Figure \ref{fig:geometric_needed_cases} visualizes the scenes where both the appearance features and the surrounding features are similar for different object instances. In this scenario, geometric cues are particularly helpful as they give penalty to the geometrically infeasible pair (\ie, false pair), hence making the overall distance of the false pair larger than that of the true pair.

\subsection{Additional Results from Structure from Motion Baseline} 

Structure from Motion(SfM) can be used to generate 3D structure from multiple views \cite{hartley2003multiple,winder2009picking}. The 3D structure can be trivially used for instance association from multiple views as pixel correspondences are known. However, an inherent limitation of SfM is that only the intersection of cameras' views can be reconstructed whereas instance association from multiple views should cover the union instead. Besides, SfM is sensitive to repetitive patterns, reflective, and textureless surfaces\cite{hirschmuller2007stereo}. We apply three state-of-the-art SfM engines, ColMap\cite{schoenberger2016sfm}, OpenMVG\cite{moulon2016openmvg}, and Theia\cite{theia-manual}, on the scenes of MessyTable. The first two are unable to reach convergence whereas Theia gives incomplete reconstruction results, shown in Figure \ref{fig:sfm_failures}.

\subsection{Visualization of SIFT Keypoints}

We visualize the keypoints detected by SIFT, as shown in Figure \ref{fig:sift}. It is clear that SIFT keypoints cluster at feature-rich regions such as edges and patterns. Texture-less instances, however, have very few keypoints. This imbalanced distribution of keypoints is likely the reason for the poor performance.




%
%
\bibliographystyle{splncs04}
\bibliography{references}

\end{document}